# Comparative Evaluation of Traditional Methods and Deep Learning for Brain Glioma Imaging. Review Paper

**Kiranmayee Janardhan[1*], Vinay Martin D'Sa Prabhu[2], T. Christy Bobby[1]**

[1]*Department of Electronics and Communications,*
*Ramaiah University of Applied Sciences*
*4th Phase, Peenya, Bengaluru, Karnataka, India*
*E-mails: kiranmayee.j@msruas.ac.in, christy.ec.et@msruas.ac.in*

[2]*Department of Radiology,*
*Ramaiah Memorial Hospital and Medical College*
*MSR Nagar, MSRIT Post, Bengaluru, Karnataka, India*
*E-mail: vinaymdprabhu@gmail.com*

[*]*Corresponding author*

**Received: August 05, 2024** **Accepted: January 06, 2025**

**Published: June 30, 2025**

***Abstract:*** *Segmentation is crucial for brain gliomas as it delineates the glioma's extent and location, aiding in precise treatment planning and monitoring, thus improving patient outcomes. Accurate segmentation ensures proper identification of the glioma's size and position, transforming images into applicable data for analysis. Classification of brain gliomas is also essential because different types require different treatment approaches. Accurately classifying brain gliomas by size, location, and aggressiveness is essential for personalized prognosis prediction, follow-up care, and monitoring disease progression, ensuring effective diagnosis, treatment, and management. In glioma research, irregular tissues are often observable, but error-free and reproducible segmentation is challenging. Many researchers have surveyed brain glioma segmentation, proposing both fully automatic and semi-automatic techniques. The adoption of these methods by radiologists depends on ease of use and supervision, with semi-automatic techniques preferred due to the need for accurate evaluations. This review evaluates effective segmentation and classification techniques post-magnetic resonance imaging acquisition, highlighting that convolutional neural network architectures outperform traditional techniques in these tasks.*

## Introduction

Medical imaging technologies, especially magnetic resonance imaging (MRI), have significantly improved the detection and prognosis of brain gliomas. The application of artificial intelligence (AI), particularly deep learning (DL) and machine learning (ML), has further enhanced the accuracy and speed of medical diagnoses [16]. Conventional glioma segmentation and classification methods, such as manual or threshold-based techniques, are being replaced by ML and DL methods due to their superior performance. These newer methods, including random forests (RFs), support vector machines (SVMs), and convolutional neural networks (CNNs), can learn complex features directly from raw MRI images without the need for manual feature engineering [4, 26].

Glioma segmentation and classification have been studied extensively using conventional ML and DL methods. Conventional methods typically involve manual or threshold-based segmentation techniques, which can be time-consuming and prone to inter-observer variability. These methods are still widely used in clinical practice but have largely been replaced by automated segmentation techniques based on ML and DL. ML techniques have shown promise in glioma segmentation and classification, including RFs, SVMs, and k-nearest neighbor (k-NN) classifiers. These methods can be trained using various features extracted from the MRI images, such as intensity, texture, and shape, to identify and classify different types of gliomas. However, these methods typically require extensive feature engineering, which can be time-consuming and limit performance.

DL methods, especially CNNs, have emerged as the most effective approach, offering improved accuracy in tasks such as glioma detection, segmentation, and grading [40]. Despite their advantages, the adoption of segmentation methods in clinical practice is slow due to the need for ease of use and user supervision rather than a lack of explainability. Explainability is a crucial factor, particularly for classification methods, to ensure clinicians can interpret and understand AI-driven decisions, addressing the challenge of their black-box nature. Techniques like saliency maps provide visual representations of input regions influencing the output, making it easier for clinicians to understand and interpret the decisions made by the classification model [40].

### *Statistics*

Gliomas are the most common malignant primary brain tumors in adults, accounting for 81% of malignant brain tumors [62] and a significant percentage of cancer-related deaths in young adults [63]. Although most gliomas are cancerous, some subtypes may not necessarily act in a cancerous manner. Despite being rare, they have a considerable effect on mortality and morbidity [41]. Glioblastoma (GBM), the most aggressive form, makes up 60-70% of malignant gliomas in the United States [62]. Gliomas also account for 2.5% of all cancer-related fatalities in people between the ages of 15 and 34, making them the third-most prevalent cause of cancer-related deaths in this age group. According to GLOBOCAN 2018 [11], the prime cause of premature death due to cancer is ranked first or second in almost 100 countries. Conforming to these proceedings, new glioma instances in India had 28.142 with a very high mortality rate (24.003 deaths).

Another analysis by the clinical cancer investigation journal in Eastern India contained 130 illustrations of gliomas. These illustrations spread between the ages of 4 to 78 years, with an average age of 42.38 years, mainly affecting the male population. According to GLOBOCAN 2018, the cancer prevalence and mortality estimates are created by the International agency for research on cancer [22]. The worldwide cancer numbers are predicted to be 28.4 million instances in 2040, which amounts to a 47% exponential incline from 2020, 64% to 95% versus transitioned (32% to 56%) nations due to globalization, owing to a growing economy demographic change.

Gliomas are categorized by the World Health Organization (WHO) into grades I to IV based on their aggressiveness, with GBM being the most aggressive (Grade IV) [49]. Manual diagnosis and delineation of gliomas are laborious and impractical in clinical settings, highlighting the need for automated segmentation tools [50]. Following standard treatment protocol, which entails surgical resection accompanied by radiation and chemotherapy, GBM patients had a mean life expectancy of 14 months and 4 months without standard treatment.

Despite various experimental therapies over the past 20 years, patient prognosis for gliomas has remained largely unchanged. Precise identification of glioma sub-region boundaries in MRI (Fig. 1) is crucial for several clinical applications, such as surgical planning and monitoring progression. However, manual diagnosis and delineation are laborious and impractical in clinical settings, underscoring the need for automated segmentation tools to streamline the process.

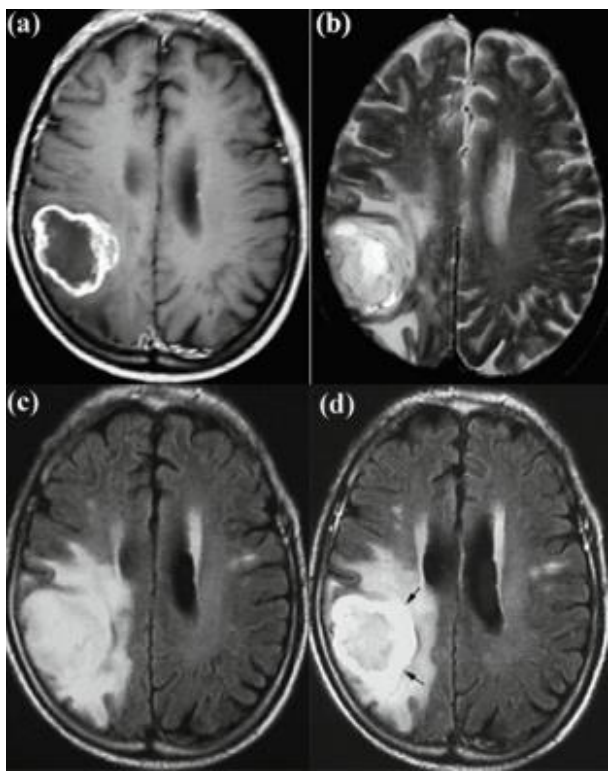

Fig. 1 Various MRI sequences of a glioblastoma multiforme, showing different sub-regions of the glioma: a) T1-weighted MRI sequence, b) T2-weighted MRI sequence, c) T1 with contrast enhancement (Gadolinium), d) FLAIR sequence. Images are taken from the MICCAI BraTS [7].

Deep learning has significantly improved the processing and analysis of biomedical images, particularly for diagnosing and treating ailments. Image segmentation, a key initial step in image processing, provides vital information about affected brain areas from MR images [55]. The segmentation of glioma images can be done manually, semi-automatically, or automatically [9]. The increasing variety of medical imaging technologies and the large volume of medical data make segmentation and classification complex [56]. Therefore, automated algorithms for detecting and classifying gliomas save time and help radiologists and clinicians make faster, more accurate, and objective diagnoses.

### *Contributions and structure*

The paper makes several key contributions to the field of glioma detection and classification. It reviews various pre-processing techniques and their impact on segmenting and classifying gliomas. It also provides a comparative analysis of different segmentation and classification techniques. Additionally, the paper examines the evaluation metrics used for identifying gliomas and the datasets these algorithms were applied, while highlighting the merits and demerits of techniques used in MR imaging for glioma detection.

The review paper aims to present a comprehensive overview of automatic glioma segmentation and classification techniques, analyzing their strengths and weaknesses. The paper discusses the datasets and pre-processing techniques, offers a critical analysis of segmentation techniques, and cover classification techniques, evaluation metrics, and methodologies of segmentation and classification.

## Datasets for glioma segmentation and classification

Fig. 2 shows the datasets that have been used by the researchers for the segmentation and classification of glioma. BraTS 2015 and 2017 are the most used datasets.

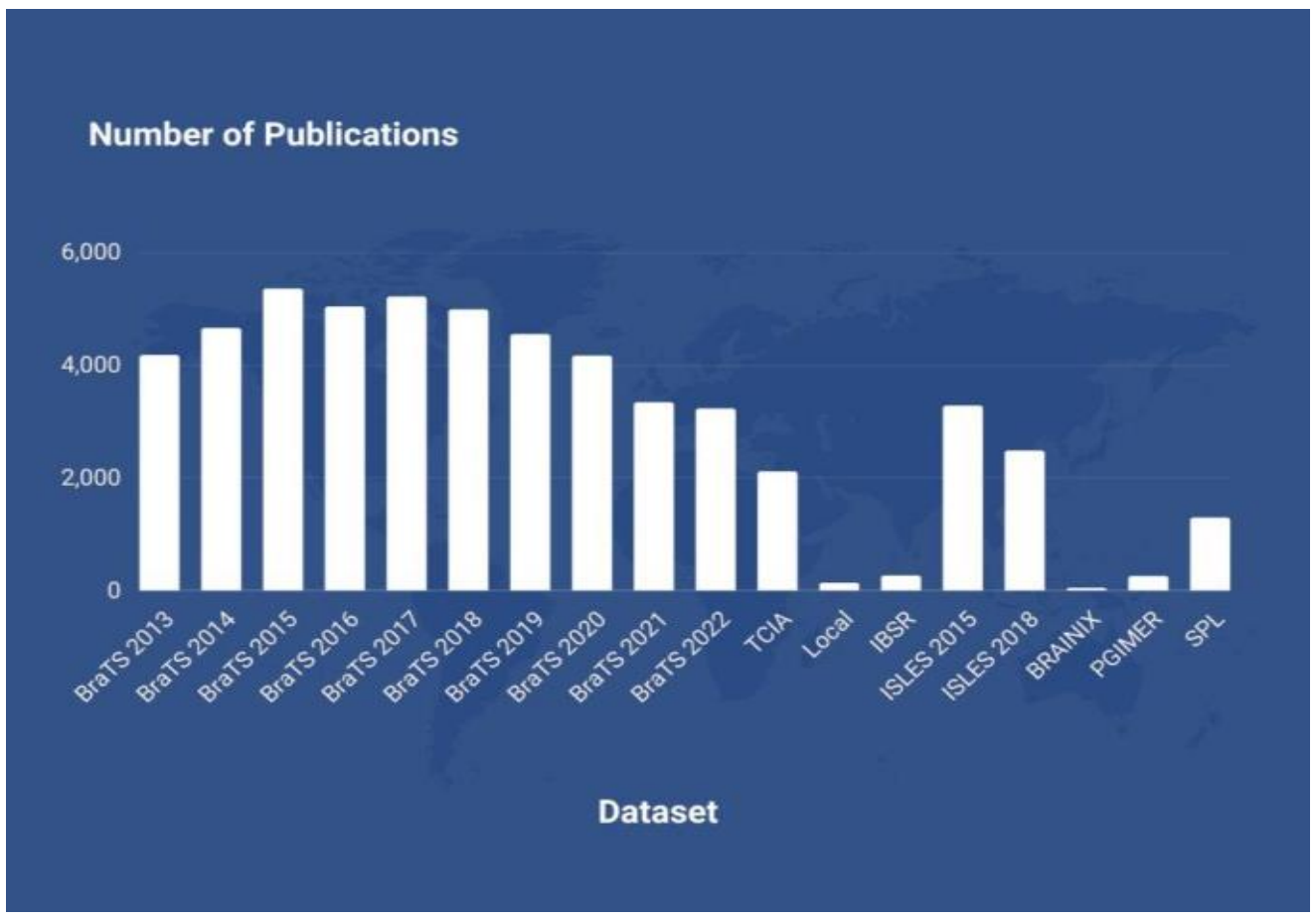

Fig. 2 Datasets used by researchers for MR brain glioma segmentation and classification

## Preprocessing techniques applied for glioma MR imaging analysis

The supreme purpose of computer vision is to utilize computers to emulate human vision. The MR image contains a high amount of information. The existing MRI scanners produce images equal to 65.535 gray levels [9]. The human eye cannot extract such data from the scans as it is not designed to determine the difference between thousands of gray levels. Consequently, a high-powered computer is the most reasonable choice for comprehending and assessing high-quality scans.

Glioma classification consists of pre-processing, image segmentation, feature extraction, feature or dimensional reduction, and classification, followed by performance analysis which results in the formulation of a diagnosis, as shown in Fig. 3, are explained below. The first stage in any research work driven by data is pre-processing. The raw image data needs to be freed of noise to guarantee uniformity in all the images present in the dataset. Image denoising, skull stripping, and intensity normalization are essential in analyzing brain images [48]. Once these operations are performed, image segmentation and extraction of features become effective.

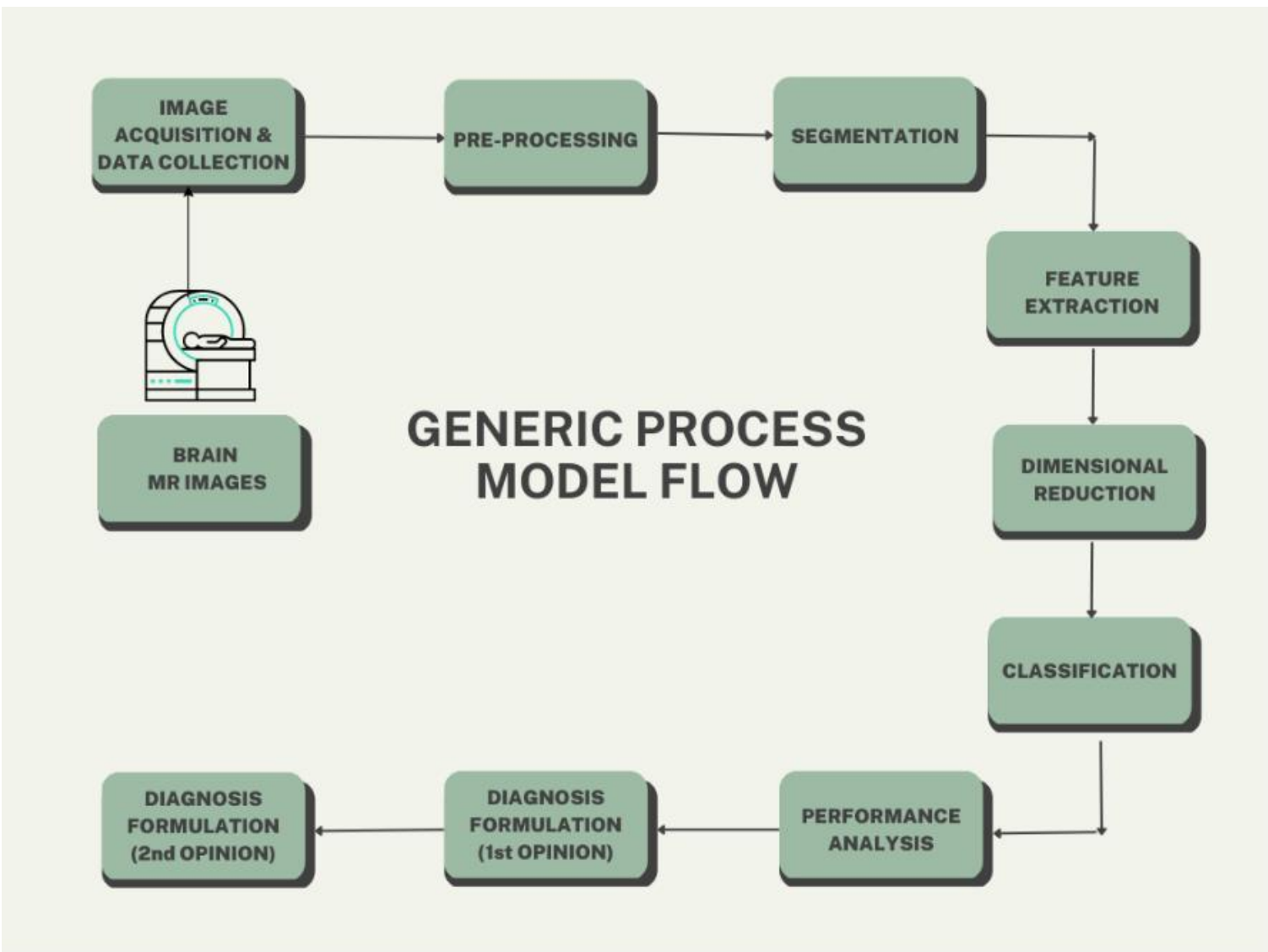

Fig. 3 Generic process flow of brain glioma segmentation and classification system for diagnosis

### *Image denoising*

Noise in digital images, caused by random signals, results in information loss and distortion. Common noise types include salt and pepper, Gaussian, speckle, Poisson, and others, arising from factors like hardware issues, light scattering, and MRI bias fields [82]. This noise can obscure lines and edges, alter the intensity [39], blur objects, and create artifacts.

Accurate image analysis of pre-and post-treatment scans is critical, yet challenging, due to patient movement. Proper alignment through image registration is necessary for assessing gliomas [58]. Noise complicates brain segmentation, making it hard to distinguish gliomas from healthy tissue. Denoising methods, such as adaptive filtering, diffusion filtering, wavelet-based methods, and non-local means (NLM), are used to enhance contrast and reduce noise [10]. Adaptive filtering is one of the standard denoising methods for magnetic resonance images during pre-processing. The diffusion filtering method preserves contours, making it robust [44]. Wavelet-based methods perform well in removing the Rician noise which reduces the image contrast as it is dependent on the signal [59]. Wavelets can preserve natural features while denoising images [61]. MRI denoising performed using the NLM method has found appreciation in [51] as it has been applied for automatic denoising. NLM can precisely fit into the noise characteristics in MR imaging [69]. Independent component analysis (ICA) is another denoising technique particularly applicable to a specific sequence of MRI [46]. Widely, ICA is applied to functional MR imaging for automatic denoising data by combining with the hierarchical fusion of classifiers [73]. Despite these techniques, some noise remains, negatively impacting glioma segmentation.

### *Skull stripping*

In image analyses of the brain, skull stripping is an essential pre-processing stage [8] as seen in skull stripping is a critical stage for segmenting gliomas [35]. Non-cerebral regions are the skull, the scalp, and the membranous meninges. The method of delineation and subtraction of the non-cerebral brain tissue is called skull stripping [20]. The precision with which skull stripping is performed influences the competency in glioma detection and pre-surgery metrics preparation. The morphometric analysis of the brain and reconstruction of the cortical surface for the quantitative study is highly dependent on the accuracy of skull stripping [75]. The comparative analysis [21] on the same has shown that it encountered numerous challenges because of the complex nature of the brain structure. The acquisition of MR image volumes could be more consistent due to the parameters of the various MR machines. The characteristics change from patient to patient, posing another challenge to skull stripping [64]. Reducing the probability of misclassifying diseased regions in the brain is better with skull removal [78]. The latest conventional and modern skull-stripping techniques have been examined to reduce misclassification [21]. Conventional skull-stripping techniques for glioma segmentation include thresholding, morphological operations, and atlas-based methods. Thresholding methods utilize a predetermined intensity threshold to distinguish between brain tissue and the background. Morphological operations employ mathematical techniques to smooth and fill holes in the thresholder image. Atlas-based methods involve registering a pre-segmented atlas image onto the patient’s image to remove non-brain tissue [48].

AI-based techniques for skull-stripping include deep learning methods such as CNN, generative adversarial networks (GANs), and U-Net. These methods have demonstrated promising results by learning features from large datasets, effectively handling complex variations in skull shape and size. Additionally, DL-based methods can be utilized for registration steps, using image similarity metrics to align the patient's image with a reference image [31].

### *Intensity normalization*

The effects cause a significant difference in the appearance of healthy and glioma tissues of the brain. Pre-processing MR image volumes involves a crucial intensity normalization stage [60]. Regarding brain segmentation, the algorithms depend on the intensity of the scanned images. Due to this, segmentation becomes a demanding task. In the algorithm developed by [18], global and local constraints were made better for the MR image volumes using their novel robust normalization technique. In [68], N4ITK algorithm has been applied to correct the intensity inhomogeneity. In a work proposed in [29], intensity normalization has been done through a CNN-based architecture.

### *Bias field correction in MRI*

Low-frequency bias field signals corrupt the MR image volumes by interfering with the high magnetic field of the MRI scanners. The volumes are further contingent on corruption caused by the noise from the patient's anatomy, which results in inhomogeneity causing variations in the intensity levels [45, 82]. The goal of correcting the bias fields is to reduce the distortion in segmentation output and to detect statistical features in MR brain images [45]. The two essential techniques applied for bias field correction are prospective and retrospective procedures [82]. The prospective mode of decreasing the bias field is modifying the MRI machines' image acquisition stage. The retrospective mode works on the acquired MR images as a post-processing technique. This method can implement bias field correction by applying filters, one of the oldest techniques. The intensity and gradient-based surface fitting techniques [54] are applied as a parametric approach along with segmentation and intensity inhomogeneity correcting histogram. In a study based on CNN approach [29], bias correction has been accomplished for glioma scans within each input channel. In the study [90], histogram matching algorithms have been employed. In the proposed work [66], deep CNNs are utilized for inhomogeneity correction with histogram normalization and transformation. In the work proposed by [43], two techniques have been explored, one uses a histogram matching and the other method is normalizing each MRI sequence with the mean cerebrospinal fluid (CSF) value.

### *Image registration*

Image registration, involving the superimposition of MR image data to align spatial, temporal, and multi-sequence parameters, is a critical aspect of medical imaging studies. This process is essential for establishing correspondence between features taken during pre-operative and post-operative treatment stages, making it a crucial pre-processing step in glioma segmentation. Both conventional and AI-based techniques can be employed for this purpose. Conventional techniques include rigid or affine transformations for basic alignment and advanced deformable registration techniques such as demons, B-spline, and diffeomorphic registration. Commonly used software tools include ANTs and Greedy, which handle multi-modal image registration effectively. Additionally, OSIRIS and OsiriX, along with ITK, ITK-snap, and 3D slicer, are widely used for their robust handling of complex 3D images and efficient visualization [37, 99].

AI-based techniques, particularly those using deep neural networks, have shown promise in learning registration parameters directly from input data, thus reducing the need for manual parameter tuning. CNN-based approaches can perform direct registration or guide conventional methods. Examples of these AI-based methods include VoxelMorph [24] and ANTsPyNet [96], both of which have demonstrated improved registration accuracy for glioma segmentation.

Despite advancements in AI, challenges remain in image registration, such as variations in patient location and organ orientation, which make it difficult to have a single technique for all problems [84]. This issue is particularly pronounced in high-grade gliomas like GBM, where rapid tumor progression necessitates precise morphological feature extraction and alignment accuracy [6]. The image registration problem is approached through four foundational frameworks: feature space registration, search space registration, search strategy-based registration, and similarity-based registration [70]. AI-based techniques have shown promising results in reducing processing time and improving registration accuracy. However, conventional techniques are still widely used in clinical practice due to their simplicity and robustness.

## Critical analysis of glioma segmentation techniques

Glioma segmentation is crucial for accurately diagnosing, outlining diseased areas, planning surgeries, and assessing treatment plans. However, the irregular and complex structures of gliomas make segmentation challenging [7]. Despite medical imaging advances, segmentation is challenging as gliomas consist of irregular and complex structures and boundaries.

### *Pixel-based segmentation*

This method, also known as threshold-based segmentation, labels pixels based on intensity values. It is computationally fast and straightforward but limited in enhancing glioma areas as it only uses intensity knowledge and ignores pixel correlation. It is often used to remove the background from MR images [26, 95].

### *Region-based segmentation*

This technique segments images into regions based on similarity criteria, starting from seed pixels and expanding to include neighboring pixels that meet the criteria. It is effective in segmenting similar areas and producing corresponding regions, but the process is recursive and stops when no more pixels can be included [38, 67, 81]. Region-based segmentation divides an image into regions ($R_1$, $R_2$, …, $R_N$) based on similarity criteria, ensuring no regions overlap and maintaining homogeneity within each region. The process stops when no more pixels meet the criteria for inclusion. However, this method's accuracy can be reduced due to edge blurring caused by the partial volume effect and its sensitivity to noise, which can result in holes in the segmented regions. Watershed and region-growing algorithms are derived from this technique [74]. Also, this technique is sensitive to noise and is prone to forming holes in the obtained region [14].

### *Edge-based segmentation*

Edge-based segmentation detects rapid intensity changes to identify borders, providing sharp edges useful for glioma detection. However, it often fails to form closed contours around gliomas and is sensitive to noise, necessitating post-processing to link edges [5]. Another disadvantage of edge segmentation techniques is that it is sensitive to image noise. The mapped image will have broken edges if its intensities change only subtly among regions [83].

### *Deformable model*

Deformable models use curves influenced by internal and external forces to fit irregular organ structures. They adapt well to the variability of human organs but face challenges in 3D segmentation due to complex control of regional modifications. Parametric models, such as active contour models (snakes), and geometric models (level sets) are used to achieve more accurate segmentation by iteratively minimizing energy functions.

The level-set deformable model's chief disadvantage is high computational complexity. The topological changes are naturally realizable, making this model highly advantageous. Earlier studies have shown that to obtain a 3D volume surface, slice-by-slice segmentation is further merged to form a continual surface. Nevertheless, the segmentation accuracy is low as the surface is discontinuous due to the missing anatomical data from the MR slices [17, 53].

### *Machine learning-based segmentation*

Machine learning methods are effective for automatic medical image segmentation, leveraging historical data for accurate patient evaluations. Image segmentation techniques can be categorized into supervised, semi-supervised, and unsupervised methods. Supervised methods use manually labelled datasets, while unsupervised methods label data automatically based on pixel intensities or textures [98]. The features used for segmentation here are based on pixel intensities or textures.

### *Fuzzy C-means*

Fuzzy C-means (FCM) clusters pixels based on attributes like intensity and texture without human intervention, making it an unsupervised method. However, it is sensitive to noise and heterogeneous intensity, leading to inaccurate segmentation [3]. FCM is sensitive to noise and heterogeneous intensity and is computationally expensive. This is its chief disadvantage, which results in inaccurate segmentation.

### *Atlas-based segmentation*

Atlas-based segmentation uses prior brain structure knowledge to aid in the new brain data segmentation, reducing computational overhead. However, it introduces bias and requires significant construction time [12]. This segmentation method is yet to be tested for generic applications as it relies on training data [57]. Additionally, bias is introduced as the technique looks for similarities in shape. Further, the main disadvantage is that the atlas needs more construction time [3].

## Deep learning segmentation – CNNs and transformers

Deep learning architectures such as CNNs and transformers are prominent in glioma segmentation. CNNs, like U-Net, excel in detecting and segmenting gliomas from MRI images [40]. Transformers, using attention mechanisms to capture spatial relationships, also show promising results, with models like vision transformer (ViT) and hybrid ViT achieving state-of-the-art performance [32].

In addition to ViT and hybrid ViT, there is the Uformer, a transformer-based architecture explicitly designed for medical image segmentation tasks. The Uformer uses a multi-scale self-attention mechanism to capture contextual information at multiple resolutions and has achieved state-of-the-art performance on the BraTS 2020 dataset. Overall, transformer-based models have shown great potential for glioma segmentation and are expected to play an increasingly important role in this field. However, as with any deep learning approach, data quality, model architecture, and hyperparameter tuning are critical factors that can significantly affect the performance of transformer-based models [94].

### *CNN-based techniques*

CNN-based models, trained on large datasets, extract features for glioma classification, significantly improving accuracy. However, they require extensive labelled data and careful handling of data variability and bias. Combining CNNs with recurrent neural networks (RNNs) and U-Net architectures can enhance accuracy by considering spatial and temporal information [13]. In this study, the authors used a deep CNN to classify gliomas based on MRI scans. They trained their model on a large dataset of over 4.000 scans from 4 different glioma types (glioblastoma, meningioma, pituitary adenoma, and brain metastasis). They achieved high accuracy in distinguishing between these glioma types.

This approach effectively improves the accuracy of glioma classification by considering the spatial relationship between different regions of the glioma. Combining CNNs, RNNs, and U-Net architectures can provide a powerful tool for glioma classification by considering both spatial and temporal information in MRI images. The work in [7] shows a CNN-based approach to glioma classification.

## Critical analysis of glioma classification techniques

In the classification stage, techniques are applied to categorize brain MRIs as either normal or abnormal. These methodologies can be classified into supervised and unsupervised methods. Supervised methodologies include SVMs, artificial neural networks (ANN), and k-NNs. Unsupervised methods include self-organizing maps (SOM) and k-means clustering. Supervised algorithms involve a training stage with precisely labelled class data, which is then applied to unlabelled data in the testing stage, generally resulting in higher accuracy than unsupervised classifiers [26]. In contrast, unsupervised techniques do not require class-labelled data as they can automatically determine the number of classes needed for classification, resolving complex problems efficiently [36].

### *Convolutional neural networks*

CNN are unique because they automatically learn image features using trainable convolutional filters, bypassing the stages described for traditional machine learning techniques. This capability allows CNNs to achieve high accuracy in various classification tasks without the need for predefined features. For instance, a recent study used a pre-trained ResNet-50 CNN architecture fine-tuned on a glioma MRI dataset to classify 6 different types of brain tumors, achieving a mean accuracy of 93.2%, outperforming traditional machine learning classifiers such as SVMs and random forests [72].

This review discusses the most utilized supervised and unsupervised classification algorithms for gliomas in MRI. Classifying different gliomas is a crucial problem as it helps develop personalized treatment strategies and predict survival outcomes. There is significant variability in glioma types, which makes this problem challenging. However, recent CNN-based techniques have shown promising results in classifying different gliomas [79].

In addition to CNN-based techniques, machine learning algorithms such as decision trees, random forests, and support vector machines have also been used for glioma classification. However, CNN-based techniques have shown superior performance in this task, likely due to their ability to extract more complex and meaningful features from the input data. Overall, classifying different gliomas using deep learning techniques holds promise for improving patient outcomes and developing personalized treatment strategies [77].

### Artificial neural network

ANN are popular for image classification due to their ability to learn from historical cases and create new rules. They use texture and intensity attributes to segment images, which are then fed to the classifier's input nodes. Mathematical computations at the hidden nodes result in the final classification at the output node. ANNs excel in complex, multivariate, non-linear environments, often performing better in noisy fields [27]. The texture and intensity attributes are employed to segment the images. These attributes are fed to the classifier's neural network input nodes, where mathematical computations are performed at the hidden nodes. The output's final node results in the image's classification [26]. No standard techniques are available for the best image classification. ANN works on complex computations of trial and error, and to work around this, convolutional neural networks (CNN) were developed in the 1980s to analyze visual data. CNN is advantageous as it can automatically detect features without human supervision [2].

The study [76] used a gray level co-occurrence matrix (GLCM), and a genetic algorithm for pre-classification, using NN classifiers to categorize MRI scans into healthy or tumorous tissue. A neural network classifier was utilized to categorize the MRI scan into healthy or tumorous tissue [86]. The algorithm for feature extraction employed cubic order differentiations. The features were selected using the rule strategy. The research employed two classifiers for the final glioma classification: feed-forward backpropagation network with k-NN. The feature extracted algorithm was obtained using the discrete wavelet transform (DWT). Reduction from 1024 to 7 features was made using the principal component analysis (PCA) [33].

Additionally, PCA was employed for extraction, and the classification of brain gliomas was performed by probabilistic neural network (PNN) [23]. A hybrid machine learning method has been developed for delineating brain gliomas in an MRI. The model comprises image data pre-processing in the suggested approach, with the median filter applied for noise reduction [19]. DWT is applied, along with its reduction by PCA. Further, the back propagation neural network (BPNN) has been employed to obtain the MR images' normality or abnormality, resulting in its classification. ANN's physical implementation is straightforward. These networks generate accurate outcomes of the generalization property, which can easily map the allocation of hybrid styles. ANNs perform perfectly in complex, problematic, multivariate non-linear environments. The performance of these NNs is better in noisy fields.

### K-nearest neighbor

The k-NN algorithm is a straightforward classification method that calculates distances between new instances and existing ones in the training set. It selects the k closest neighbors based on the distance metric, determining the majority class among those neighbors as the predicted class. While effective, especially when combined with parameter optimization techniques like cross-validation, k-NN may not be the most accurate or efficient for large and complex datasets compared to advanced techniques like CNNs and RNNs [1, 28].

### Support vector machine

SVMs, introduced by Vapnik and Cortes, are effective for non-linear decision surfaces and have applications in various fields including medical analysis [15], which have found applications in object detection, segmentation of images, voice recognition, and medical analysis [74]. SVMs have shown that they can be applied to differentiate and classify voxels into normal and abnormal tissue [42]. SVMs can classify and differentiate voxels into normal and abnormal tissue by finding a hyperplane that separates different classes in the feature space.

Studies have shown SVMs achieve high accuracy in classifying glioma grades, but they can struggle with large datasets and overlapping classes [93].

The advantage of the SVMs classifier is that it is highly effective in the higher dimension and works if the features are more significant than the training MR data. Increased accuracy is obtained when the classes are separable. SVMs are disadvantageous when the dataset is enormous as the computation time increases. The classifier does not perform well when the classes overlap.

### *Self-organizing maps*

Self-organizing map are unsupervised techniques for clustering brain MRI data into normal and abnormal categories. They provide a compact representation of data distribution and are useful for visualizing high-dimensional data. During training, model vectors are progressively adjusted, creating an organized, non-linear regression of the data. This method helps in accurately labelling gliomas by consistently grouping specific areas like gliomas and white or gray matter [34]. Thus, SOM assists in accurate glioma labelling.

### *K-means clustering*

K-means clustering is an unsupervised learning technique used for brain glioma classification. It divides the dataset into k-clusters, where each data point belongs to the cluster with the closest mean. Studies have applied k-means clustering to segment glioma regions and extract features for subsequent classification using machine learning algorithms like SVM. K-means clustering effectively identifies different tissue types in MRI, aiding in glioma classification by ensuring data points are accurately allocated within clusters [87]. They applied k-means clustering to segment the glioma region and then extracted features based on intensity values and texture features. They used these features to train a SVM classifier to distinguish between different grades of gliomas. Another study used k-means clustering to segment gliomas in MR images and classified them into low-grade and high-grade gliomas [65]. They applied k-means clustering to segment the glioma region and then extracted shape and texture features to train an SVM classifier. K-means clustering was used in both studies to segment gliomas and extract features for subsequent classification using a machine learning algorithm [97].

The classification of brain MRIs into normal and abnormal categories is essential for developing personalized treatment strategies and predicting survival outcomes for glioma patients. While traditional machine learning techniques like SVM, ANN, and k-NN have been widely used, deep learning techniques, particularly CNNs, have shown superior performance due to their ability to extract complex features automatically. Unsupervised techniques like SOM and k-means clustering also play a crucial role, particularly when dealing with high-dimensional data or when labelled data is scarce. The choice of technique depends on the specific requirements of the task, the nature of the dataset, and the desired accuracy and computational efficiency.

## Performance evaluation in image processing

Performance metrics are essential for quantitatively assessing the effectiveness of traditional and deep learning techniques in image processing. These metrics evaluate the robustness and adaptability of different methodologies, ensuring the reliability of their results. Key performance metrics include accuracy, precision, sensitivity, recall, specificity, the dice similarity coefficient (DSC), F1 score, Hausdorff distance, and the confusion matrix.

These metrics offer a comprehensive evaluation of an algorithm's performance, highlighting its strengths and weaknesses.

Accuracy measures the proportion of correctly classified instances, while precision (positive predictive value) determines how often positive predictions are correct. Sensitivity (recall or true positive rate) measures the algorithm's ability to identify positive instances correctly, and specificity (true negative rate) assesses its ability to identify negative instances accurately. The Dice similarity coefficient evaluates the overlap between predicted and ground truth segmentations, with values ranging from 0 (no overlap) to 1 (perfect overlap). The F1 score, the harmonic mean of precision and recall, is particularly useful for imbalanced datasets. The Hausdorff distance measures the maximum distance between predicted and ground truth segmentations, indicating the presence of outliers. The confusion matrix provides a detailed breakdown of true positives, false positives, true negatives and false negatives, enabling the derivation of other metrics. In the context of brain glioma segmentation and classification, these metrics are crucial for developing accurate and reliable models. Segmentation performance is often assessed using the Dice similarity coefficient and Hausdorff distance, while classification performance is evaluated using metrics like accuracy, sensitivity, specificity, precision, F1 score, and the area under the receiver operating characteristic curve (AUC-ROC). These evaluations guide the selection of the best models, ensuring improved diagnostic accuracy and aiding in the development of personalized treatment strategies.

## Discussion

Conventional image processing techniques for brain MRI segmentation utilize spatial filters, DWTs with GLCMs features, image textures, and local histograms. Machine learning techniques, such as support vector machines and random forests, are employed for pattern classification in image segmentation [26, 71, 81, 86, 89, 90, 92]. They can quickly process a large amount of data and work with different medical imaging modalities [52]. While conventional methods are reliable and have been used clinically, they rely on mathematical models and prior knowledge, which may not suit complex or heterogeneous datasets. AI-based methods like CNN and transformers can learn directly from data, making them adaptable to various datasets and achieving high accuracy. However, AI methods require substantial annotated data, and their "black box" nature can be problematic in clinical settings where interpretability is crucial.

Conventional brain glioma segmentation methods, such as manual segmentation, thresholding, region growing, and statistical methods like active contours, have been used for decades but are often time-consuming and require expert knowledge. Recent advances in deep learning, specifically CNNs and transformers, have shown promise in brain glioma segmentation and classification by automatically learning relevant features and handling large datasets efficiently. Despite their advantages, AI methods require large labelled datasets, and significant computational resources, and can struggle with data outside their training distribution.

For brain glioma segmentation and classification, combining conventional and AI-based methods can leverage the strengths of both approaches. Conventional methods offer interpretability and context, while DL methods provide high accuracy and efficiency. Supervised and unsupervised clustering methods like ANN and fuzzy C-means offer precise outcomes in heterogeneous gliomas. Various segmentation methods, including pixel-based, region-based, edge-based, and deformable models, have unique advantages and limitations, such as computational speed, handling complex structures, and sensitivity to noise.

Brain glioma segmentation involves problems with a severe difficulty level, and it is a time-absorbing task when executed with ANNs. The unsupervised methodologies of fuzzy C-means, the most favored, permit fuzzy theories to define clusters and give thoroughly precise outcomes in cases of heterogeneous gliomas. The supervised clustering method of ANN can design non-trivial distributions contributing direct practical advantages (Table 1) [92].

Table 1. Segmentation methods with advantages and disadvantages

| **Segmentation method** | **Description** | **Advantages** | **Disadvantages** |
|---|---|---|---|
| **Pixel-based (threshold-based)** | Find threshold values based on the histogram peaks of the image. | Simple and computationally fast. | Restricted applicability to enhancing glioma areas. |
| **Region-based** | Partitioning the image into homogeneous regions and topological interpretation. | Correctly segments regions with similar properties; stable results; continuous boundaries. | Partial volume effect; intensity variation can cause holes or over-segmentation; complex gradients. |
| **Edge-based** | Detection of discontinuity. | Suitable for images with better contrast between regions. | Inaccurate segmentation with objects having too many edges. |
| **Deformable models: parametric-based and level-set based** | Building models with prior knowledge of shape, orientation, location, and statistical data. | Adapts to irregular biological structures; topological changes possible. | Parametric model may converge to indefinite boundaries; long compute time. |
| **Machine learning-based** | Simulation of a learning process for decision making. | Unsupervised; converges boundaries of glioma; models non-trivial distributions; fast training time. | Long compute time; sensitive to noise. |
| **Atlas-based** | Knowledge from prior labelled training images to segment selected image. | Adaptive to variations in image intensity profiles. | Bias in segmentation output; requires more construction time. |
| **Convolutional neural networks** | Extract features using convolution kernels or filters. | Automatic feature extraction; efficient image processing; high accuracy. | Long training process; high computational requirements; difficulty with small datasets. |

Model-based algorithms, including fully automatic glioma segmentation using geometric deformable models [48] or level sets, provide accurate results but are computationally expensive. Integrating human knowledge and prior understanding of tissue characteristics can enhance segmentation outcomes. Techniques like DWTs, GLCMs, PCA, and GA are used for feature extraction and reduction, while classifiers like ANNs, k-NNs, SVMs, and SOMs perform well in MR image classification tasks [30]. Research shows that convolution neural network-based techniques produce good outcomes. CNN is different from other classifiers because it automatically learns its features from an MR image. Table 2 provides comprehensive overview of various classification methods used in the context of brain glioma analysis, detailing their respective advantages and disadvantages.

Table 2. Classification methods with advantages and disadvantages

| **Classification method** | **Description** | **Advantages** | **Disadvantages** |
|---|---|---|---|
| **Artificial neural network-based** | Based on powerful computational data models capturing complex relationships. | Straightforward implementation; accurate results; excellent performance in noisy fields. | Sophisticated network with high computational costs; complicated training sample collection; slow learning phase. |
| **K-nearest neighbor** | Classifies based on features and training data samples. | Suitable for smaller data; simple implementation. | Poor run-time performance with large datasets; potential for inaccurate results; sensitivity to redundant features. |
| **Support vector machine** | Converts data and finds optimal border between outputs. | Powerful for linear and non-linear data; significant performance on large datasets. | Accuracy affected by small datasets; difficulty in choosing optimal parameters for non-linearity separable data. |
| **Self-organizing maps** | Finds clusters in training data for classification. | Efficient for dividing M×N dimensional data into multiple segments. | Number of NN units must equal desired clusters; execution time increases with image division. |
| **K-means clustering** | Partitions dataset into k-predefined distinct non-overlapping clusters. | Fast algorithm suitable for real-time MR image data classification. | Poor outcomes with improper choice of k-value. |
| **Convolutional neural networks** | Extracts features from images for pattern recognition. | No human supervision required; automatic feature extraction; handles large datasets well. | Requires large labelled datasets; costly and time-consuming to obtain and annotate. |

In glioma imaging research, accurate cancer detection remains a primary goal. Segmentation methods help distinguish cancerous from healthy tissues using pixel intensity and threshold or region-growing algorithms. Model-based methods are employed when size and shape can identify gliomas. While segmentation techniques have evolved significantly, challenges in clarity and accountability for automated techniques must be addressed for clinical approval.

Fig. 4 illustrates the various methodologies employed in the segmentation and classification of MR brain gliomas. It highlights both conventional and AI-based approaches, showcasing their respective advantages and disadvantages. One vital objective in glioma imaging research is detecting cancer precisely. Segmentation methods conform to the attributes that differentiate cancerous tissue from healthy tissues. The segmentation process removes the uncertainty in classifying pixels within the brain's glioma region. Pixel intensity helps differentiate cancer from healthy tissues, and then threshold or region-growing algorithms are applied for segmentation. Further, some gliomas can be recognized by sizes and shapes. In such cases, model-based methods were employed for the segmentation.

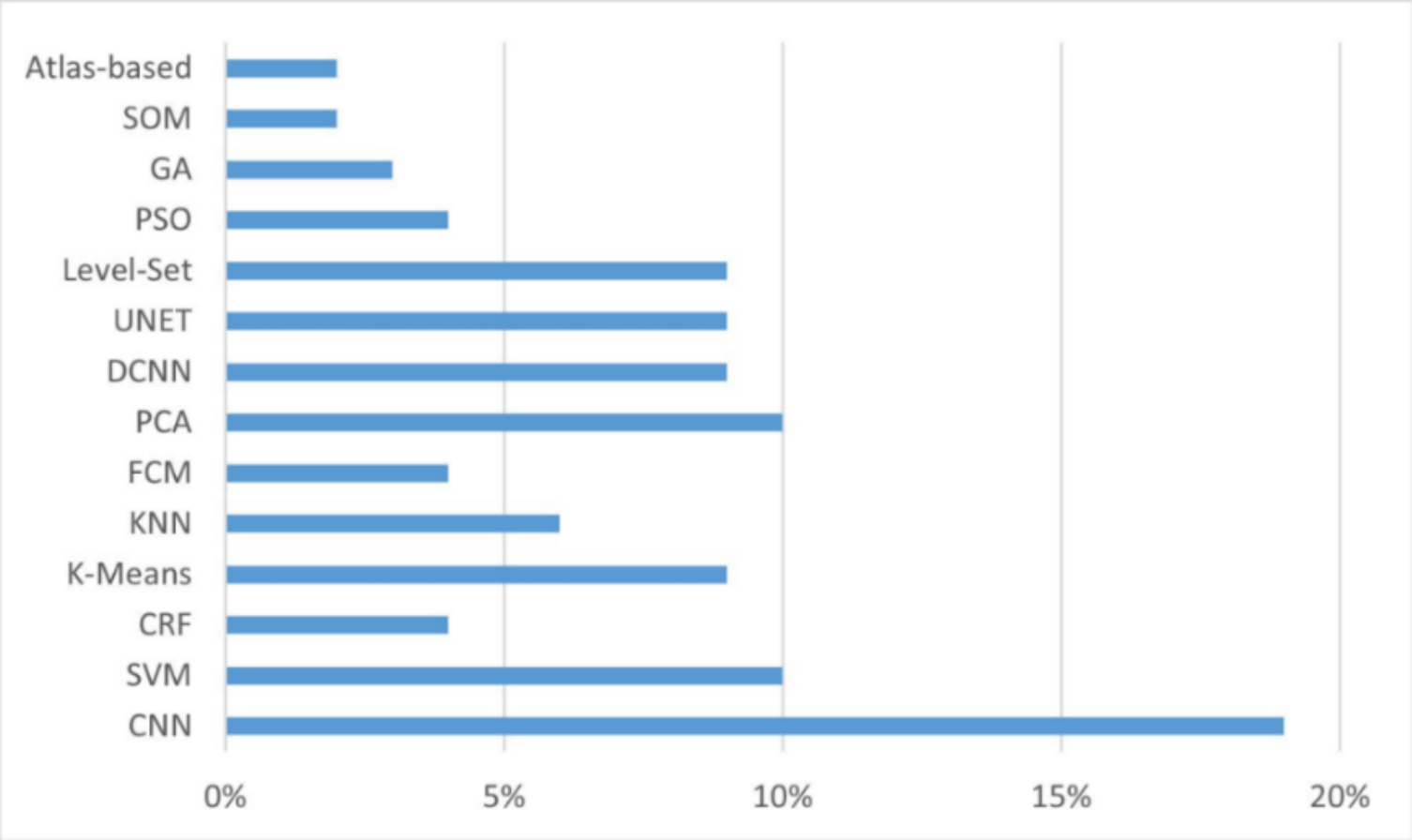

Fig. 4 Most common techniques used for MR brain gliomas segmentation and classification

Segmentation techniques are evolving and are expected to be used in clinical applications. Even though these techniques produced and accomplished exceptional feats, many challenges still need to be answered. Clarity and accountability for automated techniques for segmentation tasks are of prime importance for its approval in a clinical setting.

MRI-based brain glioma analysis faces challenges such as movement artifacts, partial volume effects, and intensity inhomogeneity. To address these issues, high-resolution MRI machines and advanced pre-filtering methodologies are employed to remove artifacts and preserve fine anatomical structures. Achieving high-resolution output requires minimizing the signal-to-noise ratio. Proper segmentation of glioma, including edema and necrotic parts, is crucial for accurate diagnosis and effective therapy protocols. Segmentation challenges arise due to variations in shape, size, and location. Therefore, enhancing the robustness, accuracy, processing time, and precision of segmentation techniques is essential for improved outcomes.

## Conclusion

The primary drive for automating brain glioma segmentation and classification lies in the need for standardized methodologies. While automated techniques show promising precision, they still require extensive validation from radiologists for routine clinical use. Two critical challenges are the substantial divergence from conventional expert practices and the current techniques' inability to support medical treatment decisions transparently and understandably. These aspects are crucial for computer-aided treatment prognosis, where clear reasoning and rationale are imperative.

This review critically examined various techniques and recent trends for classifying brain MR scan data as normal or abnormal, highlighting the use of ANNs, k-NNs, and SVMs. Traditional methods face significant limitations, such as execution time and detection accuracy, necessitating improvement. They also struggle with translating initial knowledge into probabilistic tasks and selecting highly representative features for classification. Deep learning techniques, particularly those using CNN architectures, have shown advantages in brain glioma segmentation. The effectiveness of these procedures can be validated by comparing outcomes with state-of-the-art methodologies using common brain glioma MR imaging databases.

## Acknowledgments

*We would like to express our sincere appreciation to Dr. B. S. Nanda Kumar, Associate Professor of Community Medicine at Ramaiah Medical College, Bengaluru, for his invaluable thoughts and guidance in the review of this paper. His insights have been instrumental in shaping the direction and quality of this research. This paper further contributes to the partial completion of Ph.D. research work.*

**Kiranmayee Janardhan, Ph.D. Student**
E-mail: kiranmayee.j@msruas.ac.in

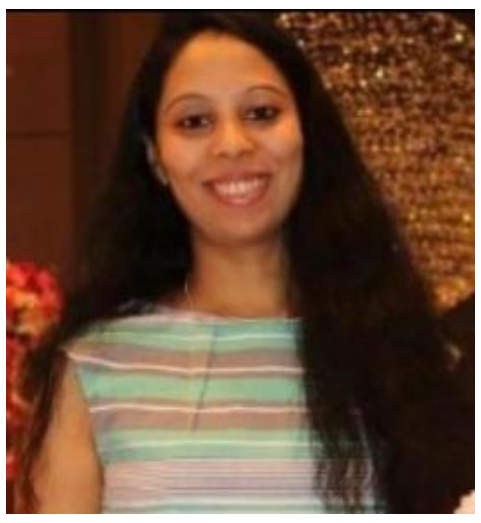

Kiranmayee Janardhan is pursuing a Ph.D. degree in Medical Imaging Using AI and Machine Learning at Ramaiah University of Applied Sciences, Bengaluru, India. Her career includes roles as a Research Scientist intern at M. S. Ramaiah Memorial Hospital, Assistant Research Scientist at LIVE 100 Hospital, and engineering roles at VMware and IBM. Her scientific interests include stem cell therapy, medical imaging, and AI-enabled diagnostics.

**Prof. Vinay Martin D'Sa Prabhu, Ph.D., M.D.**
E-mail: vinaymdprabhu@gmail.com

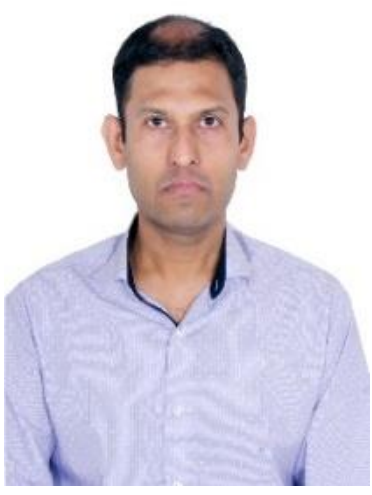

Dr. Vinay Martin D'Sa Prabhu has been a Professor in the Department of Radiodiagnosis at M. S. Ramaiah Medical College and Teaching Hospital, India since April 2013 and has served in various capacities there since 2003. He also consults at M. S. Ramaiah Memorial Hospital. He holds an MBBS, M.D. in Radiodiagnosis, and DNB from St. John's Medical College. His research includes mathematical flow modeling and adiponectin levels in diabetes.

**Assoc. Prof. T. Christy Bobby, Ph.D.**
E-mail: christy.ec.et@msruas.ac.in

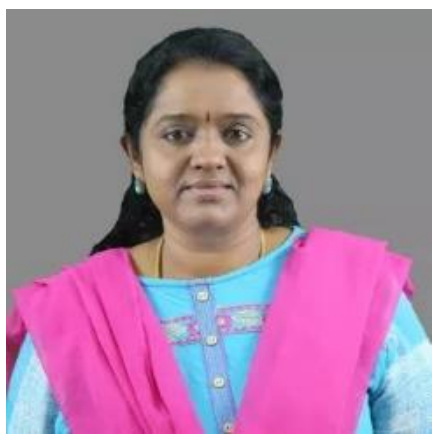

Dr. T. Christy Bobby is an Associate Professor in the Department of Electronics and Communication Engineering at Ramaiah University of Applied Sciences, Bangalore, India. With 26 years of teaching experience, Dr. Bobby specializes in biomedical signal and image processing. Her research interests include medical image processing, machine learning, and cognitive neuroscience.